\documentclass{article}

 \usepackage[preprint]{neurips_2019}
\usepackage[utf8]{inputenc} 
\usepackage[T1]{fontenc}    
\usepackage{hyperref}       
\usepackage{url}            
\usepackage{booktabs}       
\usepackage{amsfonts}       
\usepackage{nicefrac}       
\usepackage{microtype}      
\usepackage{float}
\usepackage{graphicx}
\usepackage{multirow}
\usepackage{enumitem}
\title{Towards the Use of Neural Networks for Influenza Prediction at Multiple Spatial Resolutions}

\author{
  Emily L. Aiken \\
  Harvard University\\
   Cambridge, MA 02138 \\
  \texttt{emilyaiken@berkeley.edu} \\
   \And
   Andre T. Nguyen \\
   University of Maryland \\
   Booz Allen Hamilton \\
   Columbia, MD 21044 \\
   \texttt{nguyen\_andre@bah.com} \\
   \And
   Mauricio Santillana \\
   Harvard Medical School\\
   CHIP, Boston Children's Hospital \\
   Boston, MA 02215 \\
   \texttt{msantill@g.harvard.edu}
}

\begin{document}

\maketitle

\begin{abstract}
  We introduce the use of a Gated Recurrent Unit (GRU) for influenza prediction at the state- and city-level in the US, and experiment with the inclusion of real-time flu-related Internet search  data. We find that a GRU has lower prediction error than current state-of-the-art methods for data-driven influenza prediction at time horizons of over two weeks. In contrast with other machine learning approaches, the inclusion of real-time Internet search data does not improve GRU predictions. 
\end{abstract}

\section{Introduction}

Infectious diseases affect billions of people every year and cause considerable morbidity and mortality worldwide. Influenza alone infects 35 million people in the US annually, causing 12,000-56,000 deaths \cite{cdcfludata}. Accurate real-time surveillance and forecasting of disease activity could help public health officials design timely interventions to mitigate outbreaks, but traditional healthcare-based surveillance systems are limited by inherent reporting delays. Data from the US Centers for Disease Control (CDC) on Influenza-like Illness (ILI) rates, for example, are available with approximately a 2-week delay and are frequently retrospectively revised \cite{yangargo}. Time-series machine learning methods that provide real-time estimates of disease activity at a high spatial resolution show promise for filling this temporal data gap, helping hospitals, clinics, and communities manage public health threats.

Previous computational work on improving real-time estimation and forecasting of disease activity has focused on ILI in the United States, employing methods ranging from applied machine learning and statistical modeling \cite{yangargo, gft, santillana,  argokernel, brooks, lu, wu, li, hu} to standard mechanistic epidemiological modeling \cite{wyang1, wyang2} and network approaches \cite{viboud, charu}. Many of these approaches explore the use of novel Internet-based data sources, including Google search information (GT) \cite{gft}, Twitter microblogs \cite{paul2014twitter}, and electronic health records \cite{SantillanaEHR}, to complement epidemiological data with Internet-based signals available in real time, producing accurate ``nowcasts" of influenza incidence \cite{yangargo, santillana, argokernel, lu, hu}. 

While the literature on data-driven nowcasting methods for estimating disease activity is well-developed from an epidemiological standpoint, the machine learning methods employed lag behind the state-of-the-art. The nowcasting models introduced to date mainly use variations of regularized linear regressions \cite{yangargo, lu} or, less often, random forests or support vector machines \cite{argokernel}. From a machine learning perspective, the problem of disease activity estimation is most suited to a more sophisticated and time-series specific model architecture, and thanks to the growing volume of recorded  epidemiological data, the use of recurrent neural networks (RNNs), and more specifically their variants long short-term memory (LSTM) and gated recurrent unit (GRU) networks, is increasingly feasible. 

To our knowledge, four papers to date explore neural network methods for epidemiological prediction. Wu et al. \cite{wu} apply a CNN-GRU architecture for state-level ILI estimation, Li et al. \cite{li} use a graph-structured RNN to account for networked regional disease spread, and Hu et al. \cite{hu} and Lui et al. \cite{lui} employ a fully-connected network and an LSTM, respectively, to track national ILI activity. However, these papers leave significant gaps: they do not evaluate performance with walk-forward validation (which has been shown to improve nowcasting accuracy \cite{yangargo}), with the exception of \cite{hu} and \cite{lui} they do not explore digital data sources, and, most importantly, with the exception of \cite{wu} they do not compare model performance to relevant baseline methods in the epidemiological literature. 

\subsection{Our Contributions}
Our work bridges the gap between the state-of-the-art in machine learning and in disease forecasting, comparing the performance of a GRU to previously established machine learning methods for real-time ILI estimation on the state- and city-level in the US. We find that the GRU is superior to baseline methods when there is a large reporting delay in the standard surveillance system (over two weeks). We further experiment with the inclusion of real-time Internet search-engine data from Google Trends (GT), and find that while the performance of baseline methods is improved by GT data, the GRU's performance is not. Finally, we conduct an in-depth analysis of feature importances for each model we build, as interpretability is key to effective practical use of data-driven models in public health.

\section{Methods}

\subsection{Datasets and Preprocessing}

Our state-level epidemiological dataset consists of CDC weekly ILI counts from Oct. 4, 2009 to May 14, 2017. Only 37 states without missing data are included in our analysis. Our city-level epidemiological dataset was compiled by IMS health based on weekly medical claims for 159 cities for the period Jan. 1, 2004 to July 20, 2010  \cite{viboud, charu}. We extract historical flu-related search activity from Google Trends (GT) \cite{trends} for each location for 256 key words shown in previous work to have strong correlation with ILI incidence \cite{lu}. Each dataset is split into training (first 50\%) and test (last 50\%) periods. Each time-series is normalized to the range [0, 1], where the minimum and maximum values are identified from the training period; normalization is reversed before evaluation.

\subsection{Modeling}
We construct four baseline models, built independently for each location in both datasets.
\begin{itemize}[leftmargin=*]

\item The persistence (\textbf{P}) model is the standard na\"ive baseline for time-series prediction, in which the most recently observed incidence is propagated $h$ weeks forward.

\item The linear autoregression (\textbf{AR}) uses a linear combination of $N$ autoregressive observations of ILI incidence in a given location to predict incidence at time horizon $h$ in that location. A linear autoregression incorporating synchronous Google search data, similar to the ``ARGO'' model presented in \cite{yangargo}, takes as features a linear combination of $N$ autoregressive terms and $G$ synchronous query volumes for a set of search terms from a single location.

\item The linear network autoregression (\textbf{LR}) captures spatial spread of disease, taking as features a linear combination of $N$ autoregressive terms from a set of regions $R$ available in the data set. We also implement a form of the LR for which synchronous Google search query volumes from all regions $R$ are incorporated, similar to the ``ARGO-net" model presented in \cite{lu}.

\item The Random Forest (\textbf{RF}) uses the same predictors as the LR model, but takes a nonparametric approach with a forest of 50 decision trees. As always, we experiment with inclusion of GT data. 

\end{itemize}

All models use a lookback window of $N=52$ autoregressive terms. Models that incorporate GT data use $G=256$ search terms. For models incorporating data from multiple locations, $R$ is selected via 4-fold cross validation from the set \{10, 20, 40\}, with $R$ selected independently for epidemiological time-series and GT time-series. Finally, linear regressions incorporate L1-regularization with the penalty parameter $\lambda$ chosen via 4-fold cross validation from the set $\{10^{-5}, 10^{-4}, 10^{-3}, 10^{-2}, 10^{-1}\}$, and the maximum depth of the random forest is chosen via 4-fold cross-validation from $\{2, 4, 8, 16\}$.

We implement a small Gated Recurrent Unit Neural Network (\textbf{GRU}) with a single 5-node hidden layer. Without GT data, the GRU accepts as input $N=52$ autoregressive terms from all locations in the data set and predicts incidence at the given time horizon for all locations simultaneously. When using GT data, the GRU accepts as input $N=52$ autoregressive terms and $G=10$ total synchronous Google search query volumes. We choose the $G=10$ queries with the highest correlation (in the training period) with ILI incidence in any location in the dataset. The GRU is trained on a mean-squared error objective with a dropout rate of 0.3 after the hidden layer to reduce overfitting. We use a learning rate of 0.001 for stochastic gradient descent and train the model for 1,000 epochs. 

\subsection{Training and Evaluation}

Models are trained with walk-forward validation (``dynamic training"), wherein each model is re-trained in each week with all data available in that week. In addition to eliminating forward-looking bias and allowing models to use all the available data, dynamic training has been shown in previous ILI-specific work to increase model accuracy \cite{yangargo}, and reflects how models would be used in real-world scenarios. Models are evaluated on the second half of each dataset based on the distribution of root mean squared error (RMSE) across all locations for four time horizons of prediction, $h$ = 1, 2, 4, and 8 weeks. We also conduct a set of Wilcoxon signed-rank tests to test whether the distribution of RMSE across locations is different for machine learning methods and the na\"ive persistence method. 

\section{Results}

\subsection{Accuracy}

\begin{figure}[H]
  \centering
    \includegraphics[width=1\textwidth]{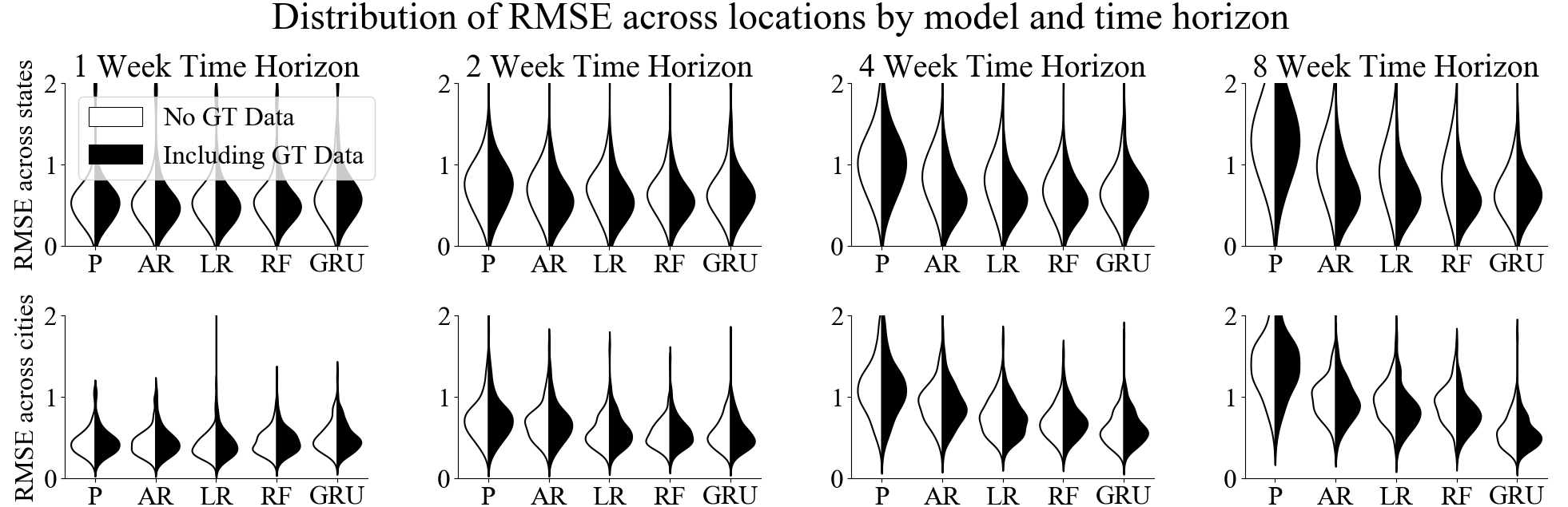}
  \caption{Summary of GRU performance in comparison to baseline models. Each violin records the distribution of prediction errors (RMSE) across locations, disaggregated by the inclusion of GT data.}
  \label{figure1}
\end{figure}

 We find that in general the GRU flu predictions have significantly lower prediction errors (RMSE) than less sophisticated machine learning models for long time horizons of prediction. Specifically, as shown in Figure \ref{figure1} and Table 1, the GRU demonstrates superior performance on 4- and 8-week time horizons for both datasets when only epidemiological data is used, and for an 8-week horizon when both epidemiological and GT data are incorporated. We observe a larger gap in accuracy between the GRU and the baseline methods on the city-level dataset. However, we find that, unlike baseline models, the GRU's performance is not improved by including real-time GT data at any time horizon.

\begin{table}[H]
\scriptsize
\centering
\begin{tabular}{llllllllll}
\hline
                                                                              &     & \multicolumn{4}{l}{No GT Data}                                                                                              & \multicolumn{4}{l}{GT Data Included}                                                                                                \\ \hline
                                                                              &     & 1 week                        & 2 weeks                      & 4 weeks                       & 8 weeks                      & 1 week                        & 2 weeks                       & 4 weeks                      & 8 weeks                     \\ \hline
\multirow{4}{*}{\begin{tabular}[c]{@{}l@{}}\rotatebox[origin=c]{90}{States}\\
\end{tabular}} & AR  & \textbf{141(1e-3)}            & \textbf{3(\textless{}e-5)}   & \textbf{9(\textless{}e-5)}    & \textbf{0(\textless{}e-5)}   & \textbf{143(2e-3)}            & \textbf{23(\textless{}e-5)}   & \textbf{8(\textless{}e-5)}   & \textbf{0(\textless{}e-5)}  \\ \cline{2-10} 
                                                                              & LR  & 340(.86)                      & \textbf{77(3e-5)}            & \textbf{4(\textless{}e-5)}    & \textbf{1(\textless{}e-5)}   & 205(.03)                      & \textbf{26(\textless{}e-5)}   & \textbf{3(\textless{}e-5)}   & \textbf{0(\textless{}e-5)}  \\ \cline{2-10} 
                                                                              & RF  & 306(.49)                      & \textbf{29(\textless{}e-5)}  & \textbf{1(\textless{}e-5)}    & \textbf{0(\textless{}e-5)}   & 268(.21)                      & \textbf{16(\textless{}e-5)}   & \textbf{1(\textless{}e-5)}   & \textbf{0(\textless{}e-5)}  \\ \cline{2-10} 
                                                                              & GRU & \textbf{100(1e-3)}            & \textbf{95(1e-4)}            & \textbf{9(\textless{}e-5)}    & \textbf{0(\textless{}e-5)}   & \textbf{114(3e-4)}            & \textbf{112(3e-4)}            & \textbf{22(\textless{}e-5)}  & \textbf{0(\textless{}e-5)}  \\ \hline
\multirow{4}{*}{\begin{tabular}[c]{@{}l@{}}\rotatebox[origin=c]{90}{Cities}\end{tabular}} & AR  & 5384(.09)                     & \textbf{4119(1e-4)}          & \textbf{2575(\textless{}e-5)} & \textbf{851(\textless{}e-5)} & 6064(.61)                     & \textbf{1907(\textless{}e-5)} & \textbf{474(\textless{}e-5)} & \textbf{14(\textless{}e-5)} \\ \cline{2-10} 
                                                                              & LR  & \textbf{2328(\textless{}e-5)} & \textbf{663(\textless{}e-5)} & \textbf{218(\textless{}e-5)}  & \textbf{13(\textless{}e-5)}  & \textbf{3996(5e-5)}           & \textbf{816(\textless{}e-5)}  & \textbf{117(\textless{}e-5)} & \textbf{6(\textless{}e-5)}  \\ \cline{2-10} 
                                                                              & RF  & 5759(.30)                     & \textbf{264(\textless{}e-5)} & \textbf{40(\textless{}e-5)}   & \textbf{2(\textless{}e-5)}   & \textbf{4159(2e-4)}           & \textbf{365(\textless{}e-5)}  & \textbf{29(\textless{}e-5)}  & \textbf{1(\textless{}e-5)}  \\ \cline{2-10} 
                                                                              & GRU & \textbf{2296(\textless{}e-5)} & \textbf{368(\textless{}e-5)} & \textbf{0(\textless{}e-5))}   & \textbf{0(\textless{}e-28)}  & \textbf{1946(\textless{}e-5)} & \textbf{333(\textless{}e-5)}  & \textbf{0(\textless{}e-5)}   & \textbf{0(\textless{}e-5)}  \\ \hline \\
\end{tabular}
\caption{Results of Wilcoxon signed-rank test comparing the distribution of RMSE for machine learning methods with the na\"ive persistence model. Test statistics ($W$), between 0 and 352 for states and 0 and 6360 for cities, indicate differences between methods (small $W$ signals a large difference), and P-values are included in parentheses with statistically significant results bolded.}
\end{table}

\subsection{Interpretability}

For interpretability purposes, we analyze feature importances across each method. Specifically, we obtain regression coefficients for each linear regression, feature importances \cite{breiman} for each random forest model, and saliency maps \cite{simonyan} for each GRU prediction, examples of which are in Figure \ref{figure2} in the appendix. We observe certain results consistent with intuitive spatial and temporal model interpretation: the most important features in linear regression and random forest models for the city-level dataset tend to be epidemiological lags and GT information from cities near to the city of prediction, and the most immediately available epidemiological information (lags 1-4) tend to be important features in linear regression and random forest models that predict under short reporting delays, while information from the previous season (lags 48-52) is more important in models that predict under long reporting delays. Similarly, saliency maps indicate that GRU attention extends much further back for neural network models that predict under a eight week reporting delay than for models working with a one week reporting delay.

\section{Discussion}
Here we introduce the use of a time-series neural network approach that improves upon the predictive accuracy of previously used machine learning methods for ILI prediction in the presence of reporting delays of over two weeks. We show that the GRU achieves superior accuracy at two spatial resolutions relevant to actionable interventions, and could therefore improve real-time tracking of ILI given the reporting delays inherent to standard healthcare-based surveillance systems. Furthermore, our results that exclude GT data indicate that under short reporting delays, the GRU could provide highly accurate forecasts of ILI activity up to 8 weeks ahead of the most recently available epidemiological report. 

We find, however, that the GRU outperforms other models only at reporting delays longer than two weeks and that the GRU is not improved by GT data. These performance differences are consistent with the trade-off between model complexity and convergence in the data-deficient scenario. The simpler models evaluated here have fewer trainable parameters than the GRU, while the GRU’s complex architecture and time series-specific structure allow it to better learn embedded patterns in historic data. The benefit of the GRU is larger when there is no external (GT) real-time data available to the model likely because the inclusion of GT data significantly increases the number of parameters. With the availability of more training data this behavior may change.

As the amount of available high spatial resolution disease-specific data grows in the field of public health, using neural network models like the one introduced here becomes increasingly feasible. Trade-offs in interpretability should be considered, however, when comparing neural networks to less complex machine learning methods. For that reason, we have presented a comprehensive feature importance analysis in this work. Note that while linear regression coefficients like the ones extracted in our analysis are highly interpretable, feature importances in a random forest model include more stochasticity and the saliency maps produced for predictions by the neural network model represent only rough approximations of model attention.

Two key limitations to this study are tuning of the neural network model and lack of access to real-time epidemiological data. First, the performance of neural network models is sensitive to several hyperparameters, including optimization choices, depth, width, and regularization. Due to computational limits, we adopt a simple GRU architecture with a single, five unit hidden layer and do not tune the model for other hyperparameters. Likely the performance of the GRU would be improved if cross-validation was used to tune key hyperparameters. Second, we have access only to final (revised) ILI data, but as noted in the introduction these data are frequently updated with post hoc revisions up until several weeks after their original release. 

There is much room for further exploration of sophisticated machine learning methods for epidemiological prediction. It would be particularly impactful to explore how well the models presented here can track other infectious diseases outside of the United States. There is also room for experimentation with other neural network model architectures. In particular, we adopt a network that is similar in size to those in past work on ILI prediction \cite{wu, li, hu}, but is very small compared to those used for other machine learning applications. We leave the exploration of deeper and wider architectures as future work.

\appendix

\section{Appendix}

\begin{figure}[H]
  \centering
    \includegraphics[width=1\textwidth]{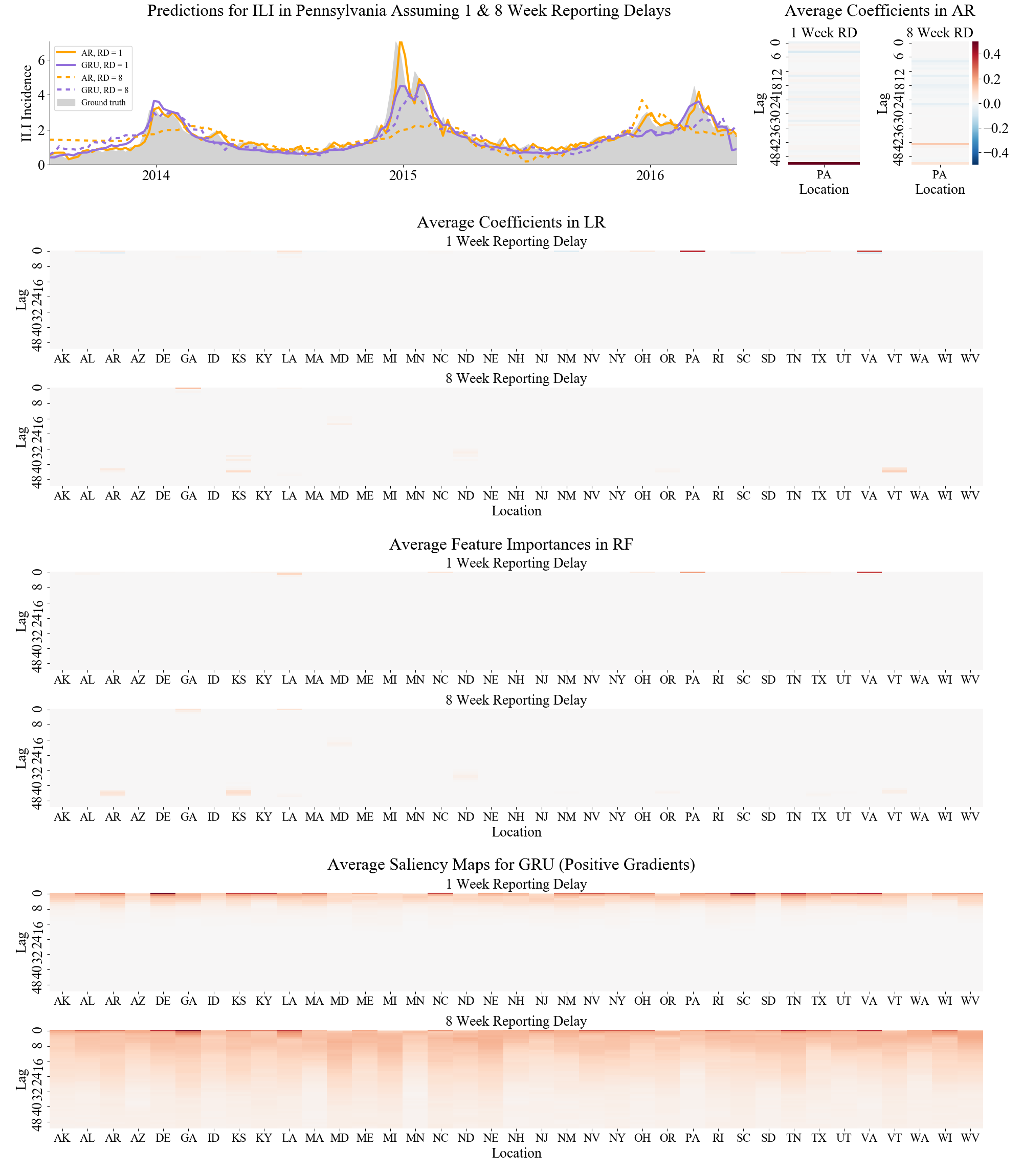}
  \caption{Example of interpretability analysis for the state of Pennsylvania. Similar analyses were performed for all states and cities. Feature importances are averaged over the entire prediction period. Note that the most important short-term predictors in the LR and RF are from Pennsylvania and nearby Virginia. Also note that GRU attention extends back much further for the 8-week prediction than for the 1-week prediction.}
  \label{figure2}
\end{figure}

\end{document}